\documentclass[final]{cvpr}

\usepackage{times}
\usepackage{epsfig}
\usepackage{graphicx}
\usepackage{amsmath}
\usepackage{amssymb}
\usepackage{booktabs}
\usepackage{multirow}
\usepackage{dcolumn}
\usepackage[utf8]{inputenc} 
\usepackage[T1]{fontenc}    
\usepackage{url}            
\usepackage{booktabs}       
\usepackage{amsfonts}       
\usepackage{nicefrac}       
\usepackage{microtype}      
\usepackage{lipsum}
\graphicspath{ {./images/} }
\usepackage{pbox}
\usepackage{epstopdf}
\usepackage{subfigure}
\usepackage{xspace}
\usepackage{comment}
\usepackage{bm}
\usepackage{bbm}
\usepackage{tabularx}
\usepackage{setspace}
\usepackage{sidecap}
\usepackage{xcolor}
\usepackage{wrapfig}
\usepackage{array}
\usepackage{xcolor}

\newcommand{\bg}{\textbf{g}}
\newcommand{\bc}{\textbf{c}}
\newcommand{\tbf}{\textbf{f}}

\usepackage[pagebackref=true,breaklinks=true,colorlinks,bookmarks=false]{hyperref}

\def\cvprPaperID{2327} 
\def\confYear{CVPR 2021}

\begin{document}

\title{P4Contrast: Contrastive Learning with Pairs of Point-Pixel Pairs for\\
RGB-D Scene Understanding}

\author{Yunze Liu$^{1,6}$\thanks{Equal contribution. Author ordering determined by coin flip.} \quad  Li Yi$ ^{2}$\footnotemark[1] \quad Shanghang Zhang$ ^{3}$ \quad Qingnan Fan$ ^{4}$ \quad Thomas Funkhouser$ ^{2}$ \quad Hao Dong$ ^{1,5,7}$\thanks{Corresponding author}\\
\\
$^1$CFCS, CS Dept., Peking University \quad
$^2$Google Research \quad
$^3$UC Berkeley \\
$^4$Stanford University \quad
$^5$AIIT, Peking University\quad
$^6$Xidian University \quad
$^7$Peng Cheng Laboratory \\
{\tt\normalsize \{liuyzchina,fqnchina\}@gmail.com, \{ericyi,tfunkhouser\}@google.com}\\
{\tt\normalsize shz@eecs.berkeley.edu, hao.dong@pku.edu.cn}
}

\maketitle

\begin{abstract}
Self-supervised representation learning is a critical problem in computer vision, as it provides a way to pretrain feature extractors on large unlabeled datasets that can be used as an initialization for more efficient and effective training on downstream tasks.  A promising approach is to use contrastive learning to learn a latent space where features 
are close for similar data samples and far apart for dissimilar ones.  This approach has demonstrated tremendous success for pretraining both image and point cloud feature extractors,
but it has been barely investigated for multi-modal RGB-D scans, especially with the goal of facilitating high-level scene understanding. 
To solve this problem, we propose contrasting ``pairs of point-pixel pairs'', where positives include pairs of RGB-D points in correspondence, and negatives include pairs where one of the two modalities has been disturbed and/or the two RGB-D points are not in correspondence. This provides extra flexibility in making hard negatives and helps networks to learn features from both modalities, not just the more discriminating one of the two.  Experiments show that this proposed approach yields better performance on three large-scale RGB-D scene understanding benchmarks (ScanNet, SUN RGB-D, and 3RScan) than previous pretraining approaches.
\end{abstract}


\begin{figure}[t]
    \centering
    \includegraphics[width=\linewidth]{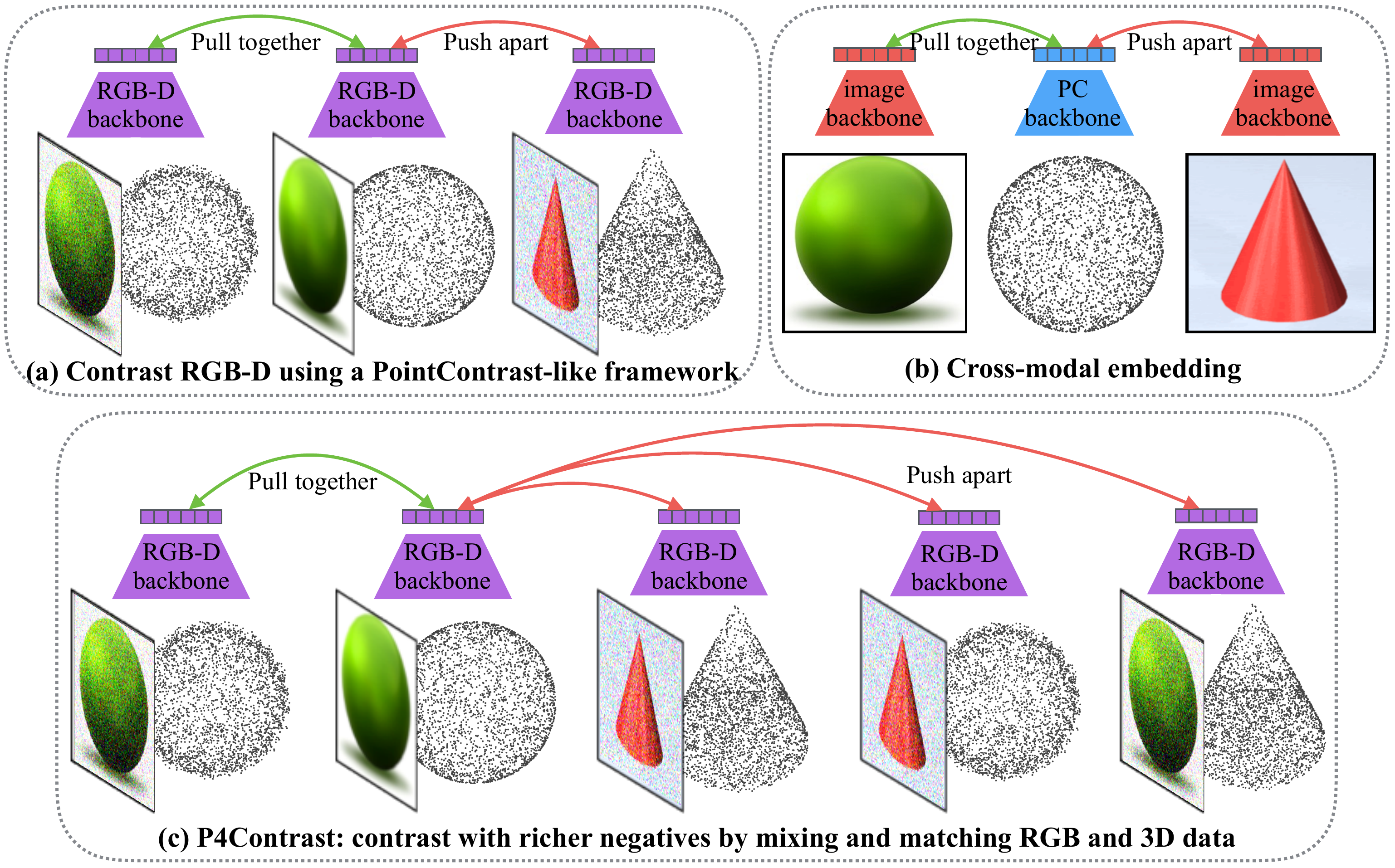}
    \caption{Different RGB-D contrastive learning frameworks.}
    \label{fig:teaser}
\end{figure}

\section{Introduction}

The goal of our work is to learn without human supervision how to extract dense features from RGB-D data that are useful for 3D scene understanding tasks such as semantic segmentation and object detection.
In the field of 2D visual understanding, a popular approach to learn such useful feature representations is contrastive learning -- where feature extractors are optimized to discriminate instances of a dataset via a contrastive loss in the latent space~\cite{hadsell2006dimensionality}. For example, SimCLR~\cite{chen2020simple} uses constrastive learning to pretrain models for image classification.
Recently, PointContrast~\cite{xie2020pointcontrast} has explored the possibility of applying self-supervised contrastive learning for high-level 3D scene understanding, with the pretext task of pulling together positive point pairs generated by transforming a point cloud with rigid transformations. The discovery is very encouraging: self-supervised pre-training on large datasets of 3D scans improves the performance of downstream 3D scene understanding tasks across different datasets.

In this paper we investigate how to make full use of contrastive learning to train dense feature extractors directly for RGB-D scans.  At first glance, the problem seems quite straight forward -- simply create point-pixel pairs via appending raw RGB values from 2D pixels to 3D points and use the same contrastive strategy as in PointContrast (Figure~\ref{fig:teaser} (a)). Surely, contrastive learning with RGB and 3D points together should produce better features than with either modality alone. 
However, we find this is not the case in practice. We observe that this simple extension fails to leverage both the color and geometry input effectively. 
Additional RGB input in the PointContrast framework brings very marginal performance improvement, sometimes even a performance degradation.

An alternative approach is to design models to extract features separately from RGB and depth channels and train them with a cross-modal contrastive loss \cite{meyerimproving,tian2019contrastive,jing2020self,gupta2016cross}, as is shown in Figure~\ref{fig:teaser} (b). These approaches force the models to extract discriminating features from 
both 2D color and 3D geometry modalities
, but it does not leverage the synergies between the two modalities to extract features better than that could be learned from either alone.  Indeed, in most cases, the model trained on the weaker modality regresses to the features learned by the stronger one, sometimes even making the features worse.

To address these problems, in this paper, we propose contrastive learning from RGB-D data with \emph{\textbf{P}airs of \textbf{P}oint-\textbf{P}ixel \textbf{P}airs}, as is shown in Figure~\ref{fig:teaser} (c). That is, we pretrain a model, called P4Contrast, that extracts deep features of point-pixel pairs and aims to pull together positive pairs where: 1) the point and pixel within each point-pixel pair come from the same RGB-D observation, and 2) the two point-pixel pairs in a pair are in correspondence with one another.  The key motivation behind this approach is that working with pairs of point-pixel pairs provides more flexibility for creating hard negatives than the previous approaches.  
Like PointContast, we can create negatives by picking pairs of point-pixel pairs which are not true correspondences.  Like cross-modal contrastive embedding, we can create negatives by replacing the RGB or 3D point within one or both of the two point-pixel pairs.  We can further create other hard negatives with combinations of these strategies.

Training a network to discriminate positive examples from all these types of negative examples encourages learning stronger features. Since the model must make a decision about whether the two components of a point-pixel pair are from the same observation, it cannot be lazy and learn strong features for only one of the two modalities. Since both components of the point-pixel pair are processed by the same network and used to discriminate correspondences between point-pixel pairs, it must learn synergistic relationships between the modalities for extracting useful features.

To evaluate this approach experimentally, we use P4Contrast to pretrain a deep backbone on a large-scale RGB-D scene dataset, ScanNet~\cite{dai2017scannet}. Then we finetune the backbone on target datasets for downstream scene understanding tasks including semantic segmentation on ScanNetV2~\cite{dai2017scannet}, semantic segmentation on 3RScan~\cite{wald2019rio} and 3D object detection on SUN RGB-D~\cite{song2015sun}.  We find that P4Contrast significantly boosts the performance of previous state-of-the-art methods for all three tasks (+\textbf{2.4} mIoU on ScanNetV2 semantic segmentation, +\textbf{4.4} mIoU on 3RScan semantic segmentation and +\textbf{2.0} mAP on SUN RGB-D 3D object detection).

Our key contributions are three-fold. First, we formulate dense RGB-D representation learning as a point-pixel pair contrastive learning problem and present a novel pretext task to encourage RGB-D information fusion. Second, we design a unified contrastive learning solution, P4Contrast, covering the pretraining objective design, the deep learning backbone construction, and data augmentation strategies. Third, we demonstrate the efficacy of P4Contrast on three large-scale RGB-D scene understanding benchmarks and we provide extensive ablation studies to validate our designs.


\section{Related Work}
\noindent\textbf{RGB-D Fusion for Semantic Scene Understanding. }
Semantic scene understanding (SSU) involves a number of fundamental vision tasks, such as object detection \cite{qi2020imvotenet,qi2019deep}, semantic segmentation \cite{hou20193d,dai20183dmv,kundu2020virtual}, and pose estimation \cite{wang2019densefusion}.
To learn better visual representations for SSU, usually both color and geometry information (RGB-D) is leveraged. The specific RGB-D data representation and fusion approaches matter for either 2.5D or 3D scene understanding tasks.
For 2.5D SSU, RGB and depth information are mostly formed as 2D images and encoded by popular 2D neural networks. The color and geometry features are combined through either early fusion~\cite{couprie2013indoor,gupta2014learning}, middle fusion~\cite{hazirbas2016fusenet,lin2017cascaded,liang2018deep,park2017rdfnet}, or late fusion~\cite{long2015fully,cheng2017locality,liu2018rgb}. For 3D SSU, some recent works lift both RGB and depth into the 3D space represented as a 6-channel point cloud (RGBXYZ), which is directly processed by the 3D network backbones for joint RGB-D feature extraction \cite{qi2017pointnet++,thomas2019kpconv,choy20194d}. The more common approach is to represent RGB as the color image and depth as the point cloud, which are processed by 2D and 3D backbones individually first for later RGB-D feature fusion \cite{wang2019densefusion,hou20193d,qi2020imvotenet,dai20183dmv,huang2019texturenet}. In this paper, we aim for 3D SSU, and argue that RGB and depth information should be complementary to each other regardless of the specific data representation and the accompanying network backbones. Thus  we propose to fuse color and geometry information in both 2D and 3D backbones to learn better visual representations for scene semantics.

\vspace{2mm}\noindent\textbf{Contrastive Self-supervised Representation Learning. }
Contrastive learning (CL)~\cite{bachman2019learning, he2020momentum, oord2018representation,tian2019contrastive, misra2020self} is a representative method of Self-supervised learning (SSL) that has gained increasing attention and demonstrated promising results~\cite{caron2020unsupervised, garnot2020metric,grill2020bootstrap, chen2020simple, chen2020improved, chen2020big}. 
Most CL methods are instance-level, aiming to learn an embedding space where samples from the same instance are pulled closer and samples from different instances are pushed apart~\cite{wu2018unsupervised, he2020momentum}. 
Recently, Chen et al. proposed a simple framework for CL (SimCLR)~\cite{chen2020simple} with larger batch sizes and extensive data augmentation, achieving comparable results with supervised learning. 
SimCLR requires a large minibatch size to achieve superior performance, which is computationally prohibitive. MoCo~\cite{he2019moco,chen2020mocov2} improves the efficiency of CL by storing representations using a queue that is independent of minibatch size. More recently, CL based on prototypes has shown promising performance~\cite{li2020prototypical, asano2020self, caron2020unsupervised, garnot2020metric}.
While CL has been actively explored for 2D image understanding, few works have been done for 3D scene understanding. Very recently, PointContrast~\cite{xie2020pointcontrast} initiates the efforts through presenting a CL framework to learn dense point cloud features.
However, it is not specifically designed for RGB-D scans. When color information is available, it simply treats the RGB values as additional features of 3D points, failing to leverage both the color and geometry input effectively. 


\vspace{2mm}\noindent\textbf{Self-Supervised Multimodal Learning.}
Extensive studies on multimodal learning are dedicated to modelling the multiple modalities and their complex interactions with the aim at leveraging complementary information present in multimodal data and yielding more robust predictions~\cite{wang2014multi, gao2019rgb, liang2019strong, xu2017multi, chen2018progressively, liang2018multimodal}. 
The multimodal feature fusion can be typically categorized as early~\cite{d2015review}, late~\cite{morvant2014majority,shutova2016black}, and hybrid fusion~\cite{baltruvsaitis2018multimodal}. 
Recently, several methods have been developed to learn cross-modal embedding in a self-supervised way~\cite{tian2019contrastive,jing2020self,meyerimproving,jiao2020self,shi2020contrastive,cheng2020look,sayed2018cross,mahendran2018cross}. 
Tian et al. proposed multiview CL to learn a representation that aims to maximize mutual information between different views of the same scene but is otherwise compact~\cite{tian2019contrastive}. Mahendran et al. learned pixels embeddings to enable the similarity between their embeddings matches that between their optical flow vectors~\cite{mahendran2018cross}.
Existing contrastive RGB-D representation learning methods mostly focus on cross-modal embedding~\cite{meyerimproving,tian2019contrastive,jing2020self,gupta2016cross}. Their feature extractors still consume single modality input but emphasize more on the correlated feature between RGB images and 3D point clouds. Those existing works are mostly tested on single objects or for low-level tasks such as registration.

\section{Method}
In this section, we introduce our self-supervised contrastive learning pipeline. Since our work is closely related to PointContrast~\cite{xie2020pointcontrast}, we first provide a brief review in Section~\ref{sec:pointcontrast}. To cope with the multimodal input in RGB-D scans and learn representations to better fuse color and geometry information, we then introduce our novel self-supervised pretraining solution, P4Contrast, in Section~\ref{sec:pretexttask}, which covers the pretext task formulation as well as the loss design. We detail the network architecture that can leverage both 2D and 3D context during feature learning in Section~\ref{sec:multicontext}, we discuss our data augmentation strategy in Section~\ref{sec:data_aug} and talk about the dataset we use for pretraining in Section~\ref{sec:dataset}. 

\subsection{PointContrast Revisited}
\label{sec:pointcontrast}
PointContrast~\cite{xie2020pointcontrast} is a contrastive learning framework that can learn descriptive dense and local geometric features on 3D point cloud. Through simple self-supervised pretraining on a large set of real 3D scenes, useful representations could be learned to boost the performance of a range of high-level 3D scene understanding tasks, such as 3D semantic segmentation and 3D object detection. A key observation is that high-level semantic scene understanding tasks require not only global but also local geometric features, making directly contrasting point cloud instances and extracting global representations insufficient. PointContrast instead contrasts pairs of points. The pretext task requires minimizing the feature distance between corresponding points from two views of a point cloud and maximizing the feature distance between unmatched points. 

PointContrast uses the Sparse Residual U-Net (SR-UNet)~\cite{choy20194d} as the network backbone that is designed for 3D point cloud processing. It makes full use of geometry information without emphasizing the role of color information in RGB-D scene understanding. In fact, under the PointContrast framework, simply treating colors as additional channels of 3D points does not bring much performance boost. The above observation motivates our work, P4Contrast, which learns to better fuse color and geometry signals in a novel contrastive learning framework.


\subsection{Contrasting Pairs of Point-Pixel Pairs as a Pretext Task}
\label{sec:pretexttask}

We assume our input of commodity RGB-D scans is a reconstructed 3D point cloud $\bg\in\mathbb{R}^{N\times3}$ and a set of RGB images $\{\bc^{(m)}\}_{m=1}^M$ that represent the geometry and color information respectively, where $M$ is the number of input images. Without loss of generalizability, we omit the image index $m$ in the following context.
We also assume the correspondences between points in $\bg$ and pixels in $\{\bc^{(m)}\}$ are established through projection, which allows us to easily bring the geometry and color signals into the same domain either in the form of colored point cloud or RGB-D image. Here we represent a corresponding point-pixel pair as ($\bg_i$, $\bc_i$), which denotes the combination of geometry and color information regardless of the underlying domain.
Different to PointContrast that learns a per-point representation, we learn to densely extract per-pair representations that encode both pixels and points with a contrastive framework.
Our goal is to pretrain a deep net backbone to effectively fuse the color and geometry signals and facilitate downstream RGB-D scene understanding tasks.


\noindent\textbf{Contrasting pairs of point-pixel pairs}
The common contrastive learning frameworks leverage anchor, positive and negative samples. The objective is to bring closer anchor-positive pairs and pull apart anchor-negative pairs.
Following this generic idea, PointConstrast treats points as the sample representation and contrasts pairs of points. In this paper, we propose to replace the pure geometric points with point-pixel pairs as the sample representation which brings the geometric and color knowledge together, and hence contrasts \textit{pairs of point-pixel pairs}.

To be specific, given an RGB-D scan ($\bg$, $\bc$), we firstly apply data transformations to obtain two versions ($\bg^1$, $\bc^1$) and ($\bg^2$, $\bc^2$) of it. This provides anchor point-pixel pairs $(\bg_i^1,\bc_i^1)$ and positive point-pixel pairs $(\bg_i^2,\bc_i^2)$. The negative point-pixel pairs can be naively computed via using all the non-positive ones $\{(\bg_j^2, \bc_j^2)| j\neq i\}$. Then the goal of contrastive objective is to extract local features that support correspondence computation.
However, we argue this is not an ideal solution while handling the multi-modal RGB-D inputs, since either geometric or color feature can be already strong enough to support the correspondence computation, and hence the feature extraction network could safely ignore the other information (color/geometry) without violating the objective much, which phenomenon can be observed in many different scenarios~\cite{xie2020pointcontrast,meyerimproving,choy20194d}.

\begin{figure}[t]
    \centering
    \includegraphics[width=\linewidth]{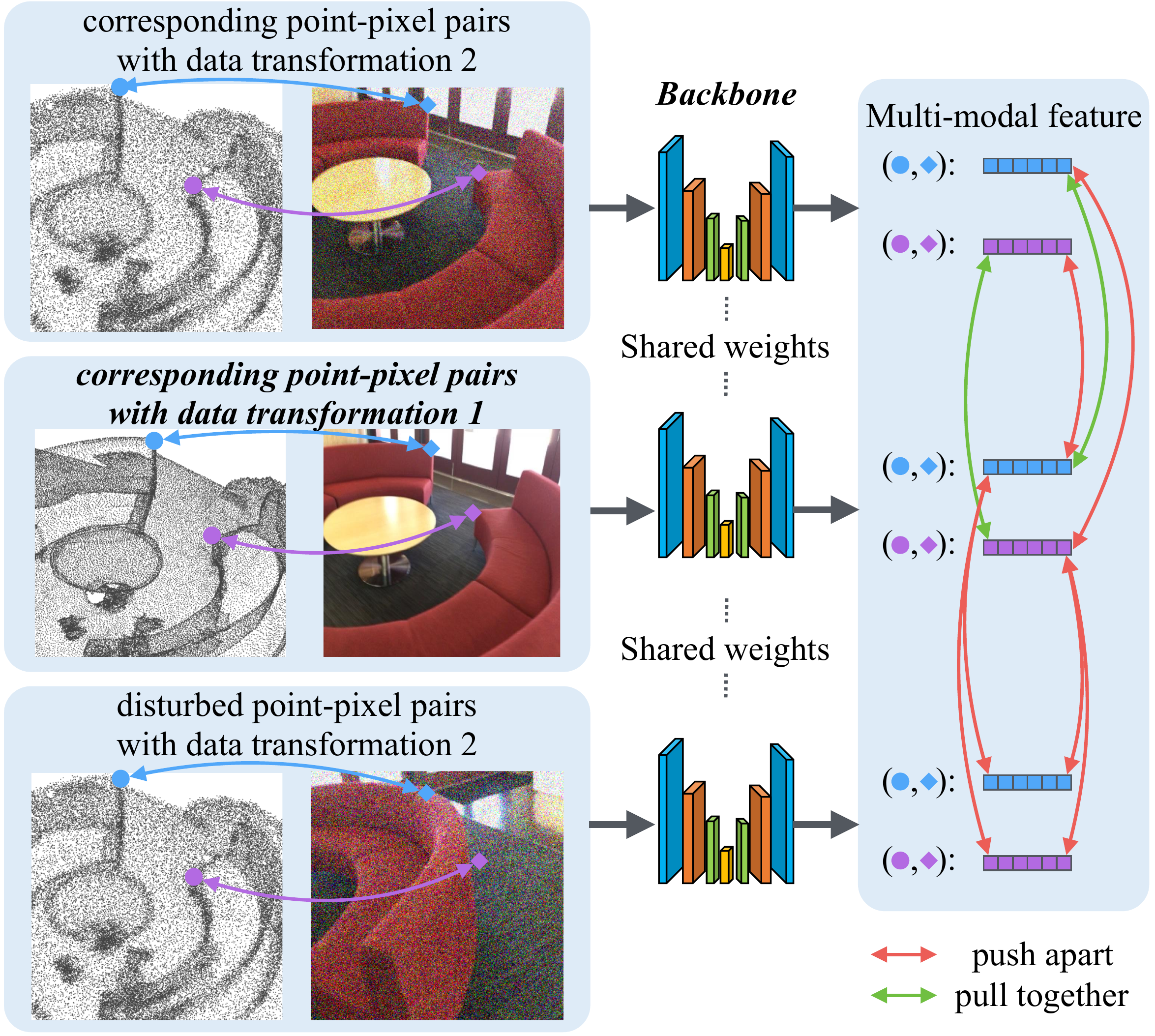}
    \caption{The pretext task contrasting pairs of point-pixel pairs.}
    \label{fig:p4contrast}
    \vspace{-10pt}
\end{figure}

\noindent\textbf{Disturbed point-pixel pairing}
The above observation indicates that a contrastive objective forcing the feature extractor to pay attention to both color and geometry input is required for better representation learning. Toward this end, we introduce \textit{disturbed point-pixel pairing} to prepare the negative samples. For an anchor point-pixel pair $(\bg_i^1, \bc_i^1)$, in addition to its unmatched point-pixel pairs $\{(\bg_j^2, \bc_j^2)|j\neq i\}$, we further break the correspondences between  points and pixels to generate a new set of negative samples $\{(\bg_j^2, \bc_{d(j)}^2)\}$, where $d(\cdot)$ is a disturbing function satisfying $d(j)\neq j$. 
In this case, both the correspondence between geometric positions of two transformed scenes and the binding between geometric and color features are broken to form our negative point-pixel pair.
Notice we do not exclude partially negative pairs $(\bg_i^2, \bc_{d(i)}^2)$ from the negative set, meaning a point-pixel pair is only positive when both the point and pixel parts are positive. This strategy forces the underlying network to extract meaningful features from both geometry and color information. We train the feature extractor through optimizing a contrastive loss over the point-pixel pairs. To be specific, we minimize the distance between anchor and positive point-pixel pairs while maximizing the distance between anchor and negative point-pixel pairs. We visualize this idea in Figure~\ref{fig:p4contrast}.

\noindent\textbf{Hardness of partially negative point-pixel pairs}
Disturbing point-pixel pairing breaks the correspondences between points and pixels and introduces negative samples that could force the feature extractor to jointly depict geometry and color. Among the disturbed pairs, we find the hardness of partially negative pairs playing a big role on the representation learning process.
Here we consider an anchor pair $(\bg_i^1, \bc_i^1)$ corresponding to point $\bg_i^1$ and a negative pair $(\bg_i^2, \bc_{d(i)}^2)$ whose point part is positive while the pixel part corresponds to another point $\bg_{d(i)}^1$. In order to discriminate $(\bg_i^1, \bc_i^1)$ from $(\bg_i^2, \bc_{d(i)}^2)$, the feature representation for the anchor pair can not ignore the color information $\bc_i^1$.

Due to spatial smoothness, the difficulty of discriminating $(\bg_i^1, \bc_i^1)$ from $(\bg_i^2, \bc_{d(i)}^2)$ usually increases as point $\bg_i^1$ and $\bg_{d(i)}^1$ get closer in the 3D space. Therefore we intuitively define the hardness of a partially negative point-pixel pair $(\bg_i^2, \bc_{d(i)}^2)$ regarding an anchor pair $(\bg_i^1, \bc_i^1)$ as $\frac{1}{||\bg_i^1-\bg_{d(i)}^1||_2}$.
The partially negative pairs are especially challenging yet important negative samples. Making them too hard, say re-pairing a point with the colored pixel from a nearby point, could confuse the feature extractor. On the other hand, the point-pixel incompatibility within a very easy pair may soften the impact on the feature extractor to discriminate the color and geometry information, therefore contributing less to the representation learning process. These require a strategy properly controlling the hardness of partially negative pairs during learning.

\begin{figure}[t]
    \centering
    \includegraphics[width=\linewidth]{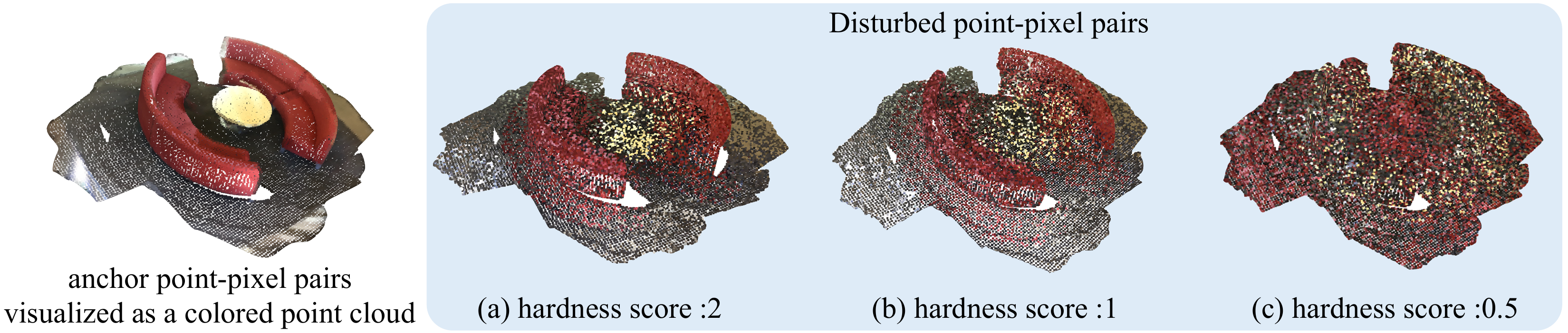}
    \caption{Disturbed point-pixel pairs with varying hardness.}
    \label{fig:hardness}
    \vspace{-12pt}
\end{figure}

\noindent\textbf{Progressive hardness increasing} We adopt a strategy of progressive hardness increasing while preparing the partially negative pairs for contrastive learning. To be specific, at training iteration $k$, we sample a random index $j$ satisfying $\frac{1}{||\bg_i^1-\bg_j^1||_2}>\text{clip}(h(k), \epsilon)$ and $j\neq i$, where $h(\cdot)$ is a linearly growing function, $\text{clip}(\cdot, \epsilon)$ clips an input value making sure the output is smaller than $\epsilon$. We re-pair $\bg_i^2$ with $\bc_j^2$ as a partially negative pair of the anchor pair $(\bg_i^1, \bc_i^1)$. Intuitively speaking, the lower bound for the hardness of partially negative pairs is gradually increasing along the learning process. We visualize the partially negative pairs with different hardness as colored point clouds in Figure~\ref{fig:hardness}. It can be seen that partially negative pairs with higher hardness have a higher correlation with the anchor point-pixel pairs, increasing the difficulty of discriminating anchor point-pixel pairs.

\noindent\textbf{Loss design}
Inspired by the PointInfoNCE loss proposed in~\cite{xie2020pointcontrast}, we use a PairInfoNCE loss to contrast different point-pixel pairs. The loss encourages an anchor point-pixel pair to be similar to its positive pair and dissimilar to many negative pairs. Specifically, our loss $\mathcal{L}_c$ is defined as follows:

\begin{equation*}
    \mathcal{L}_c=-\sum\limits_{i}\text{log}\frac{\text{exp}(\tbf_{ii}^1\cdot\tbf_{ii}^2/\tau)}{\sum\limits_{j\neq i}{\text{exp}(\tbf_{ii}^1\cdot \tbf_{jj}^2/\tau)}+\sum\limits_k{\text{exp}(\tbf_{ii}^1\cdot \tbf_{kd(k)}^2/\tau)}}
\end{equation*}

\noindent where $\tbf_{ij}^v$ denotes the extracted feature of point-pixel pair $(\bg_i^v, \bc_j^v), v=1,2$ and $d(\cdot)$ is the disturbing function breaking the correspondences between points and pixels. Notice the negative samples contain both disturbed point-pixel pairs and undisturbed ones. $d(\cdot)$ is designed to increase the hardness of partially negative pairs progressively during training.

\subsection{Learning Backbone Capturing 2D-3D Context}
\label{sec:multicontext}

\begin{figure}[t]
    \centering
    \includegraphics[width=\linewidth]{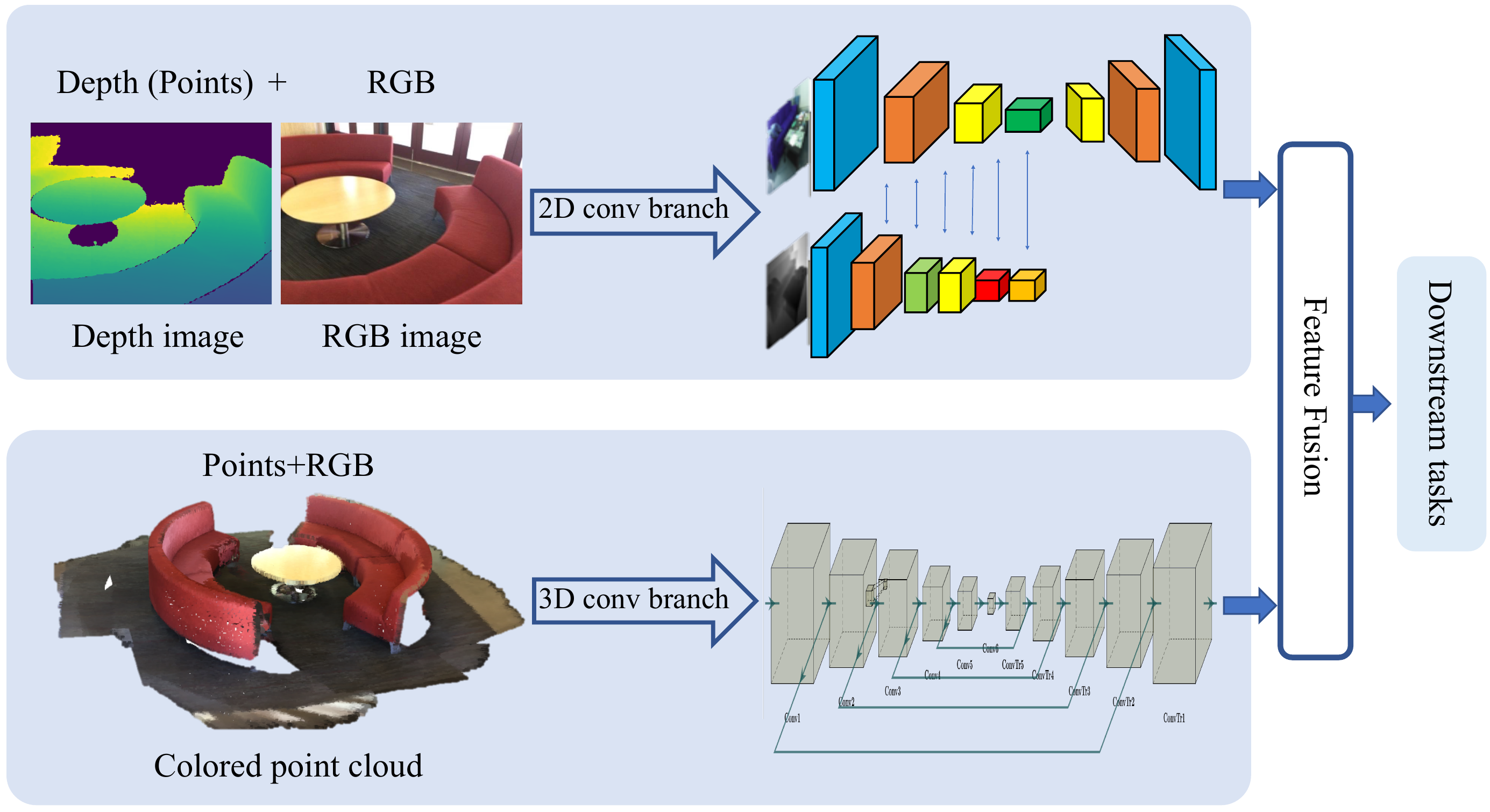}
    \caption{2D-3D context contrastive learning backbone.}
    \label{fig:backbone}
    \vspace{-12pt}
\end{figure}

To learn the dense RGB-D feature representations, a deep learning backbone is required to consume the input point-pixel pairs ($\bg_i$, $\bc_i$). Note the point and pixel is the representation of geometry and color information, which can be brought into the same domain in the form of either 3D colored point cloud or 2D RGB-D image through camera projection. Therefore the point-pixel pair ($\bg_i$, $\bc_i$) can be encoded by both the popular 2D and 3D convolutional network backbones.
2D convolutional backbones leverage 2D context where the distance of point-pixel pairs is determined by the color pixel $\bc_i$ while in 3D convolutional backbones the distance of point-pixel pairs is computed using the geometric point $\bg_i$. The difference of the underlying metric spaces influences how features get aggregated. We conject that 2D and 3D context should be complementary to each other and a good deep learning backbone should not restrict itself to either one. We therefore present our 2D-3D context contrastive learning backbone, which combines a 3D and a 2D backbone as is shown in Figure~\ref{fig:backbone}.

\noindent\textbf{Contrasting with 3D context}
We use a Sparse Residual U-Net (SR-UNet) as our 3D backbone which is designed in~\cite{choy20194d} and also leveraged by PointContrast. The backbone treats a 3D point cloud as a set of sparse voxels and leverages sparse 3D convolutions in a 34-layer U-Net architecture~\cite{ronneberger2015u}. Each input point is equipped with a 6-channel feature including its 3D location and RGB color. The SR-UNet consumes the 6-channel input points and densely generates per-point feature representations.

\noindent\textbf{Contrasting with 2D context}
We use FuseNet~\cite{hazirbas2016fusenet} as our 2D backbone, which is originally designed for RGB-D image semantic segmentation. FuseNet contains a RGB network and a depth network, both with 2D convolutional encoder-decoder structures. The RGB network and the depth network first extract their own features at the front end of FuseNet. Then sparse fusion is used to fuse their features and the output is a dense feature map.

\noindent\textbf{Contrasting with 2D-3D context}
Our overall backbone simply combines the 3D SR-UNet and the 2D FuseNet by concatenating their corresponding output features. Notice our backbone combines both early fusion happening inside 2D/3D backbones, and late fusion merging the outputs of 2D and 3D backbones, therefore could be categorized as a hybrid fusion backbone.

\subsection{Data augmentation}
\label{sec:data_aug}

\begin{figure}[t]
    \centering
    \includegraphics[width=\linewidth]{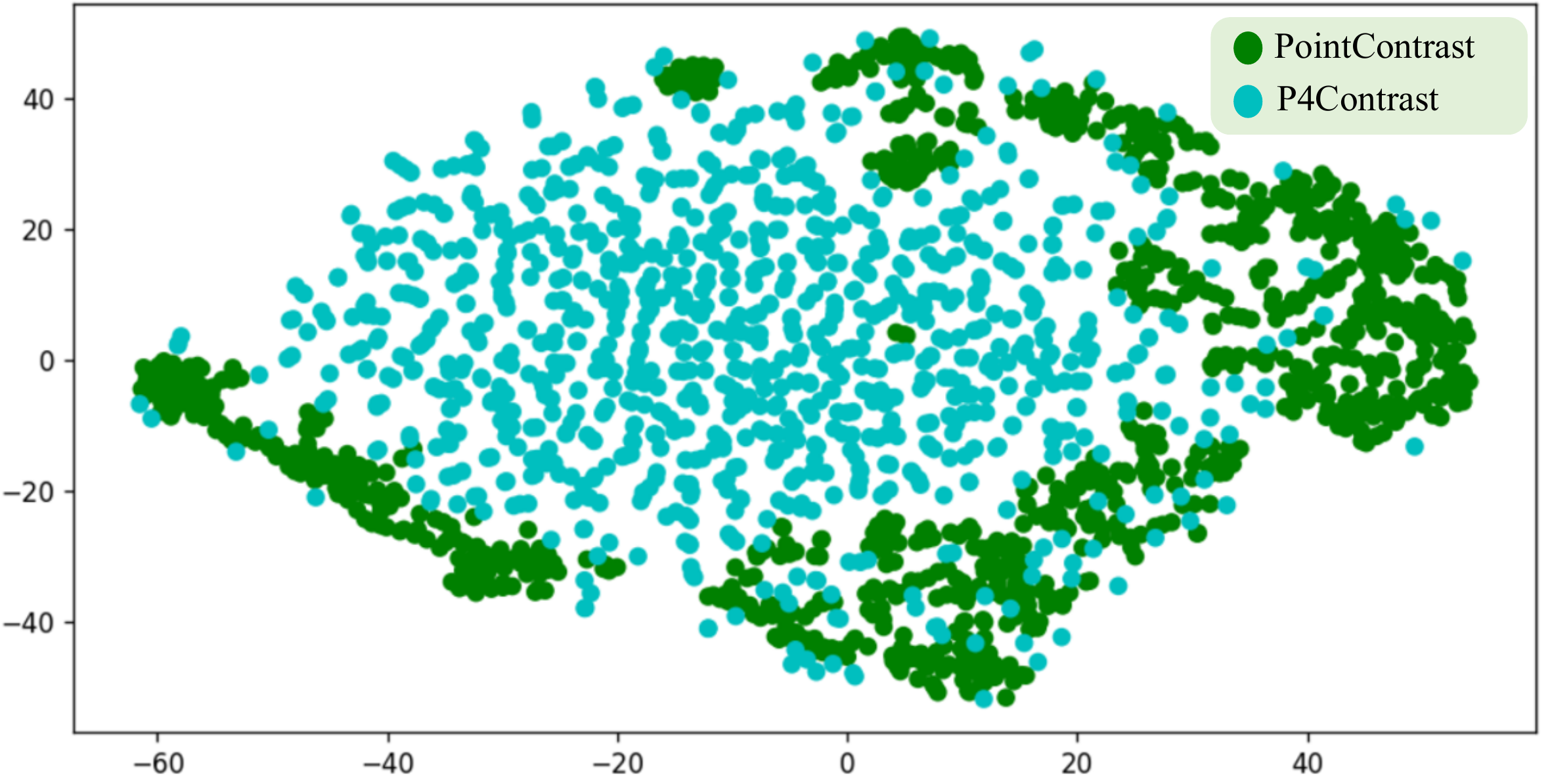}
    \caption{The t-SNE visualization of the pretrained features with data augmentation strategies from PointContrast and P4Contrast. Features tend to collapse in space when data augmentation consists of random rigid transformation plus scaling. We only use point jittering as the augmentation strategy in P4Contrast, producing more uniformly distributed features.}
    \label{fig:tsne}
    \vspace{-12pt}
\end{figure}

Data augmentation is an important component in contrastive learning which allows generating anchor, positive and negative examples and extracting representations invariant to certain noise in the input. As contrastive learning becomes more and more popular, what are suitable data augmentations has also attracted many interests ~\cite{xiao2020should}, mostly restricted to 2D images though. PointContrast as the first work leveraging contrastive learning for dense and local 3D feature learning,  proposes a 3D data augmentation strategy containing two steps: rendering a scene point cloud into a depth view and applying random rigid transformations plus scaling afterwards. We discover through experiments that this augmentation strategy is quite vulnerable to \emph{mode collapsing} (features collapsed in space) with the SR-UNet backbone as is shown in Figure~\ref{fig:tsne}. This will restrict the benefits of pretraining for downstream semantic understanding tasks. After experimenting with different 3D data augmentation strategies, we end up using only point jittering~\cite{qi2017pointnet++} to generate different versions of a scene point cloud.
Though losing rotation and scale invariance, we find this allows very robust training against \emph{mode collapsing}, producing more uniformly distributed features which has been argued to be beneficial to downstream tasks previously~\cite{bojanowski2017unsupervised}. As for RGB image views, we add Gaussian noise to an RGB image to generate different versions of it~\cite{chen2020simple}, which allows us to easily maintain the point-pixel correspondences.

\subsection{Dataset for Pretraining}
\label{sec:dataset}

Following PointContrast, we use ScanNet~\cite{dai2017scannet} as the dataset for pretraining. ScanNet is a large-scale indoor scene dataset containing \textasciitilde1500 RGB-D scans. Each scan contains a reconstructed point cloud and a sequence of RGB images with the ground truth point-pixel correspondences known. We firstly pretrain the SR-UNet and FuseNet backbone separately and then fuse their output features together. While training SR-UNet, we use colored point cloud to represent a ScanNet scene. As described in Section~\ref{sec:data_aug}, we jitter point positions and colors with Gaussian noise to generate anchor, positive and negative examples. While training FuseNet, we sample paired RGB and depth images from a scene. Again we augment RGB and depth images with Gaussian noise.

\subsection{Implementation Details}
\par For ScanNet and 3RScan semantic segmentation task, we train the model with one V100 GPU for 60,000 iterations. Batch size is 16. We use SGD+momentum optimizer with an initial learning rate 0.8. We use polynomial learning rate scheduler with a power factor of 0.9. We set the weight decay as 0.0001 and the voxel size is 2.5cm. Above hyper parameters are  mostly consistent with PointContrast.
\par In the experiment of SUN RGB-D 3D object detection, our settings are consistent with ImVoteNet. We train the model on one 2080Ti GPU for 180 epochs. The initial learning rate is 0.001 and the tower weights are 0.3,0.3,0.4 respectively. We sample 20,000 points from each scene and the voxel size is 5cm.

\section{Experiment}
Following PointContrast, we adopt a supervised fine-tuning strategy to evaluate how well the learned representations could transfer to downstream tasks. Specifically, we first pretrain our 2D-3D backbone on ScanNet dataset with the proposed pretraining objective and data augmentation strategies. Then we use the pretrained weights as initialization and further refine them for target downstream tasks. The performance gain will be a good indicator measuring the quality of learned features. In this section, we cover three popular RGB-D scene understanding tasks: semantic segmentation on ScanNetV2~\cite{dai2017scannet}, 3D object detection on SUN RGB-D~\cite{song2015sun} and semantic segmentation on 3RScan~\cite{wald2019rio} in Section~\ref{sec:scannet}, \ref{sec:sunrgbd} and \ref{sec:3rscan} respectively. We also demonstrate the benefits of representation learning when the training set is small in Section~\ref{sec:small_trainingset}. In addition, we provide extensive ablation studies to validate our design choices in Section~\ref{sec:ablation}.

\subsection{Fine-tuning on ScanNet semantic segmentation}
\label{sec:scannet}

We first conduct experiments to see whether self-supervised pretraining on ScanNet dataset can help semantic scene understanding on its own. With P4Contrast we can explore the synergetic signals between color and geometry and learn representations robust to noise and potentially different from what obtained with a direct supervised training. Therefore, there is a good reason to believe our self-supervised pretraining should improve semantic segmentation even when the source dataset for pretraining and target dataset for evaluation are the same.
We follow the official data split of ScanNet with 1,201 training scenes and 312 validation scenes. We pretrain on both the training and validation set and conduct supervised finetuning using only the training set. We evaluate the semantic segmentation on the validation set covering all the 20 semantic categories considered in the official benchmark. Mean IoU(mIoU) is used as the evaluation metric. We compare our method with baselines trained from scratch without pretraining, as well as PointContrast~\cite{xie2020pointcontrast} which treats point-pixel pairs as individual ``points''. To better understand the contribution from the pretraining objective, we factorize the backbone difference and provide a variation of our approach using the SR-UNet 3D backbone only (P4Contrast(3D Context)). We compare our full approach, P4Contrast(2D-3D Context), with all above methods and report the results in Table~\ref{tab:scannet_semseg}. Notice we also highlight the input leveraged by different methods where ``Geo'' denotes point cloud only and ``Geo+RGB'' adds 2D color images.

We notice in~\cite{xie2020pointcontrast}, comparisons with baseline methods on ScanNetV2 semantic segmentation task do not factorize the backbone difference. In~\cite{xie2020pointcontrast}, the from scratch training baseline leverages a SR-UNet with 3D convolution kernel of size 5 while PointContrast leverages a SR-UNet with 3D convolution kernel of size 3, making the comparisons unfair. To factorize the influence from backbones as well as some other hyperparameters, we use the officially released code from PointContrast, build the 3D backbone of P4Contrast upon that while keeping the hyperparameters unchanged as much as possible. In addition, for a more complete and fair comparison, we re-run the PointContrast code as well as the from scratch training baseline using backbones with 3D convolution kernels of both size 3 and 5. We use PointContrast\footnotemark[1] to denote results from re-running the officially released code. Possibly due to some hyperparameter change, we find the released code of PointContrast fails to reproduce results reported in the paper. Since our framework is largely built upon the released version PointContrast\footnotemark[1], we can still fairly compare our results with PointContrast\footnotemark[1].

\begin{table}[h]
\centering
\caption{Semantic segmentation results on ScanNet validation set. $\text{mIoU}_{\text{K5}}$ and $\text{mIoU}_{\text{K3}}$ are mIoU obtained with backbones of convolution kernal size 5 and 3 respectively.}
\label{tab:scannet_semseg}
\newcolumntype{Y}{>{\centering\arraybackslash}X}
{
\setlength{\tabcolsep}{0.2em}
\begin{tabularx}{\columnwidth}{>{\centering} m{0.47\columnwidth}|>{\centering} m{0.2\columnwidth}|Y|Y}
\toprule
   Methods & Input & $\text{mIoU}_{\text{K5}}$ & $\text{mIoU}_{\text{K3}}$\\
    \hline 
   \multirow{2}{*}{Train from scratch}&Geo &71.3&72.1 \\
   &Geo+RGB &72.2&73.2\\ 
    \hline
  PointContrast~\cite{xie2020pointcontrast} &Geo+RGB&$\backslash$&74.1\\
   \hline
  \multirow{2}{*}{PointContrast\footnotemark[1]} &Geo &72.4&73.2 \\
   &Geo+RGB &72.7&73.8 \\
   \hline
  P4Contrast(3D context) &Geo+RGB &73.6&74.3 \\
   \hline
  P4Contrast(2D-3D context) &Geo+RGB &\textbf{74.6} &\textbf{75.0}\\
\bottomrule
\end{tabularx}
}
\end{table}

We have several discoveries from Table~\ref{tab:scannet_semseg}. 
Using backbones with 3D convolution kernels of size 5 as an example, PointContrast has improved the segmentation mIoU by $1.1\%$ over the from-scratch-training baseline when only point cloud is used as input. When color information is also available, the improvement drops to $0.5\%$. Appending color values as additional features to 3D points does not introduce much benefit in the PointContrast framework, with only a $0.3\%$ mIoU improvement. The comparison between P4Contrast(3D context) and PointContrast\footnotemark[1] directly validate the efficacy of contrasting pairs of point-pixel pairs v.s contrasting ``points'' containing color features since they use the same SR-UNet backbone. The improvement from $72.7\%$ to $73.6\%$ confirms that with our novel pretraining objective design,  we can better leverage the color information to boost the segmentation performance on RGB-D scenes. With our full approach whose backbone leverages both 2D and 3D context information, we can achieve a $2.4\%$ mIoU improvement over the baseline trained from scratch. Similar observations apply to the setting using backbones with 3D convolution kernels of size 3, where the segmentation mIoU is $0.4\%-1.1\%$ higher than with kernel size 5.

\subsection{Fine-tuning on SUN RGB-D 3D object detection}
\label{sec:sunrgbd}

We then focus on another popular RGB-D scene understanding task, 3D object detection. We also switch to a different target dataset, SUN RGB-D~\cite{song2015sun}, to test how pretraining on ScanNet scenes could benefit high-level understanding of single-view RGB-D scans. SUN RGB-D contains a training set with \textasciitilde5K single-view RGB-D scans and a test set with \textasciitilde5K scans. The scans are annotated with amodal 3D oriented bounding boxes for objects from 37 categories. Following the standard protocol~\cite{song2016deep}, we only evaluate and report the results on 10 most common object categories. Our 2D-3D backbone supports pixel/point-wise label prediction but does not directly output 3D bounding boxes, therefore we need to modify our backbone for detection. In analogy to PointContrast modifying and finetuning on VoteNet~\cite{qi2019deep}, we adapt our backbone based on the current state of the art RGB-D 3D object detection architecture, ImVoteNet~\cite{qi2020imvotenet}. ImVoteNet backbone leverages PointNet++~\cite{qi2017pointnet++} to process depth point cloud and uses Faster R-CNN~\cite{ren2015faster} pretrained on COCO $train2017$ dataset~\cite{lin2014microsoft} to process input images, which aligns well with our 3D convolutional backbone and 2D convolutional backbone. To also leverage the power of supervised pretraining from COCO, we only pretrain the PointNet++ backbone on ScanNet in this case but with both RGB and point cloud inputs using the objective from P4Contrast. We compare with several baseline methods including VoteNet~\cite{qi2019deep}, PointContrast~\cite{xie2020pointcontrast} and ImVoteNet~\cite{qi2020imvotenet} in Table~\ref{tab:sunrgbd_detect}. We use mAP@0.25 as our evaluation metric.

\begin{table}[h]
\centering
\caption{SUN RGB-D 3D object detection results.}
\label{tab:sunrgbd_detect}
\newcolumntype{Y}{>{\centering\arraybackslash}X}
{
\setlength{\tabcolsep}{0.2em}
\begin{tabularx}{\columnwidth}{Y|Y|Y}
\toprule
   Methods & Input  & mAP@0.25\\
    \hline
  ImVoteNet~\cite{qi2020imvotenet} &Geo+RGB  &63.4 \\
   \hline\hline
   \multirow{3}{*}{VoteNet}&Geo  &57.7 \\
   &+RGB  &56.3 \\
   &+region feature &59.6 \\
    \hline
  \multirow{2}{*}{PointContrast}&Geo  &57.5 \\
  &Geo+RGB  &61.8 \\
   \hline
  ImVoteNet\footnotemark[2] &Geo+RGB  &61.5 \\
   \hline
  P4Contrast &Geo+RGB  &\textbf{63.5} \\
\bottomrule
\end{tabularx}
}
\end{table}

It is worth mentioning that we use the official code from the authors of ImVoteNet as our detection backbone. After re-running the code, we observe a $1.9\%$ mAP drop compared with the numbers reported in~\cite{qi2020imvotenet}. We confirm with the authors and find this is due to changes to the codebase and hyperparameters in the released version. For a fair comparison, we report these numbers as ImVoteNet\footnotemark[2] in Table~\ref{tab:sunrgbd_detect} since our method pretrains ImVoteNet\footnotemark[2] with P4Contrast.
VoteNet and PointContrast do not fully explore how to leverage images to boost the performance of RGB-D 3D object detection, their performance is outperformed by ImVoteNet\footnotemark[2] by an obvious margin. Using PointContrast to fine-tune ImVoteNet, we observe very limited improvement from 61.5 to 61.8 which still lagsbehind our proposed P4Contrast by a significant margin. After pretrained with P4Contrast on ScanNet, we obtain a $2.0\%$ gain on mAP@0.25. This again validates that P4Contrast could encourage effective fusion of color and geometry signals and our pretrained representations generalizes well across datasets.

\subsection{Fine-tuning on 3RScan Semantic Segmentation}
\label{sec:3rscan}
To demonstrate the efficacy of P4Contrast on more benchmarks, we conduct another supervised fine-tuning experiment on 3RScan~\cite{wald2019rio} dataset focusing on the semantic segmentation task. 
3RScan is a large-scale real indoor scene dataset which features 1482 3D reconstructions of 478 naturally changing indoor environments, containing a training set with 385 scans and a test set with 47 scans. In comparison to scans from ScanNet which are captured with a Structure Sensor, scans in 3RScan are captured with Google Tango with different noise patterns and scan qualities. Such differences bring extra challenges while transferring the pretrained representation. As we will show, P4Contrast still does a good job to improve the downstream 3RScan semantic segmentation task. We evaluate and report the results on the predefined 27 categories in 3RScan dataset, and adopt mean IoU (mIoU) as the evaluation metric. The numerical results are reported in Table \ref{tab:3RScan_semseg}.


\begin{table}[h]
\centering
\caption{Semantic segmentation results on 3RScan validation set.}
\label{tab:3RScan_semseg}
\begin{tabular}{c|c|c} 
    \toprule
    Methods & Input & $\text{mIoU}_{\text{K5}}$\\
    \midrule 
    Train from scratch & Geo+RGB & 37.3\\
    PointContrast & Geo+RGB & 38.8\\
    P4Contrast (3D context) & Geo+RGB & 40.8\\
    P4Contrast (2D-3D context) & Geo+RGB & \textbf{41.7}\\
    \bottomrule
\end{tabular}
\end{table}

Compared to PointContrast, P4Contrast (3D context) improves the mIoU from 38.8 to 40.8. As these two approaches share the same network backbone and only differs in their strategy to process the input color and geometry information, the improvement validates the effectiveness of contrasting pairs of point-pixel pairs. After applying the full pipeline of P4Contrast, our performance is further upgraded to 41.7, and achieves 4.4 mIoU improvement over the train-from-scratch baseline.


\subsection{Fine-tuning on Small Training Set}
\label{sec:small_trainingset}

Representation learning has exhibited powerful transfer capability that enables pretraining on a large-scale dataset and fine-tuning on a different target set. Good representations should be able to bring even larger performance gains when the target fine-tuning dataset is small in scale. To validate this, we reduce the amount of labeled data while fine-tuning for ScanNet semantic segmentation task. Different from Section~\ref{sec:scannet}, we only finetune the pre-trained network on 10\% of the training set but still test on the whole validation set. Again, we compare with the train from scratch baseline as well as the PointContrast method using mIoU as the evaluation metric. We show the experimental results in Table \ref{tab:small_semseg}. As expected, the full pipeline of P4Contrast is still able to yield better performance over all the other competitors. Moreover, we observe an mIoU improvement of 4.5 over the train from scratch baseline when only using 10\% of the training set, which almost doubles 2.4 mIoU improvement when using 100\% of the training set.



\begin{table}[h]
\centering
\caption{Semantic segmentation on ScanNet with small training set.}
\label{tab:small_semseg}
\begin{tabular}{c|c|c} 
    \toprule
    Methods & Input & $\text{mIoU}_{\text{K5}}$\\
    \midrule 
    Train from scratch & Geo+RGB & 59.9\\
    PointContrast & Geo+RGB & 62.3\\
    P4Contrast (3D context) & Geo+RGB & 63.9\\
    P4Contrast (2D-3D context) & Geo+RGB & \textbf{64.4}\\
    \bottomrule
\end{tabular}
\vspace{-10pt}
\end{table}


\subsection{Analysis Experiments and Discussions}
\label{sec:ablation}

In this section, we provide more analysis to provide an in-depth understanding of our framework. We use ScanNetV2 semantic segmentation as the target downstream task throughout the section and mIoU is used as the evaluation metric. We use SR-UNet as our 3D convolutional backbone and FuseNet as our 2D convolutional backbone. The 3D convolution kernels used in SR-UNet is of size 5.

\noindent\textbf{Multi-context or single-context}
PointContrast restrict itself to a 3D convolutional backbone which can only capture 3D context while loses the dense 2D patterns in RGB-D input. We instead propose to contrast with both 2D and 3D context through a backbone combining 2D convolution from FuseNet and 3D convolution from SR-UNet. To validate this design choice, we experiment with three different backbones: FuseNet with just 2D context; SR-UNet with just 3D context; and FuseNet+SR-UNet combining 2D and 3D context. In addition to P4Contrast, we also adapt PointContrast by changing the backbone it uses. In all the experiments we feed both RGB and point cloud inputs to the backbone (colored point cloud for 3D convolution and RGB-D image for 2D convolution). The results are shown in Table~\ref{tab:multi_context}. For both PointContrast-like framework and P4Contrast, using 2D-3D context has the best segmentation performance. P4Contrast outperforms PointContrast-like framework consistently by over $1\%$ mIoU using all three backbones, indicating its better capability of fusing color and geometry input.

\begin{table}[h]
\centering
\caption{ScanNet semantic segmentation after pretrained with different context. We report mIoU on the ScanNet validation set.}
\label{tab:multi_context}
\newcolumntype{Y}{>{\centering\arraybackslash}X}
{
\setlength{\tabcolsep}{0.2em}
\begin{tabularx}{\columnwidth}{>{\centering} m{0.25\columnwidth}|>{\centering} m{0.2\columnwidth}|>{\centering} m{0.2\columnwidth}|Y}
\toprule
   & 2D context & 3D context & 2D-3D context\\
    \hline
   PointContrast &58.9 &72.7 &73.2 \\
    \hline
   P4Contrast &60.1 &73.6 &\textbf{74.6} \\
\bottomrule
\end{tabularx}
}
\end{table}

\noindent\textbf{The influence of pair hardness}
We introduce disturbed point-pixel pairing in P4Contrast, which constructs partially negative point-pixel pairs. These challenging negative pairs encourage the learned representation to focus on both geometry and color inputs. However, directly pretraining the backbone with very hard partially negative pairs could confuse the network and get training stuck. On the other hand, always training with easy partially negative pairs will not be able to squeeze the power of disturbed point-pixel pairing fully. To demonstrate this, we experiment with three different hardness scheduling strategies. One is our proposed strategy, progressive hardness increasing, which raises the hardness lower bound of partially negative pairs following a clipped linear function $\text{clip}(h(k),\epsilon)$ as training iteration number $k$ increases, where $\text{clip}(\cdot, \epsilon)$ sets any values larger than $\epsilon$ to $\epsilon$. Another is to always use $h(0)$ as the hardness lower bound. The last is to always use $\epsilon$ as the hardness lower bound. We use ``Progressive'', ``Easy'' and ``Hard'' to represent these three strategies respectively in Table~\ref{tab:pair_hardness}. As can be seen, our progressive hardness increasing strategy leads to the best performance.

\begin{table}[h]
\centering
\caption{The influence of pair hardness on P4Contrast.}
\label{tab:pair_hardness}
\newcolumntype{Y}{>{\centering\arraybackslash}X}
{
\setlength{\tabcolsep}{0.2em}
\begin{tabularx}{\columnwidth}{Y|Y|Y}
\toprule
   Easy & Hard & Progressive\\
    \hline
    73.8&72.0 &\textbf{74.6} \\
\bottomrule
\end{tabularx}
}
\end{table}

\noindent\textbf{What data augmentation to use}
We have mentioned that the data augmentation strategy leveraged in PointContrast is quite vulnerable to mode collapsing in our experiment as is shown in Figure~\ref{fig:tsne}. That augmentation strategy mainly consists of random rigid transformation to the input point cloud. To figure out the best augmentation strategy for P4Contrast, we conduct a range of controlled experiments. Our base strategy transforms an input RGB-D scan by adding Gaussian noise to both the color image and the 3D point cloud. We test different variations including: no RGB image augmentation, add point cloud rotation augmentation, add point cloud translation augmentation, random point cloud scaling, add point cloud flipping augmentation and multi-view rendering. The results are shown in Table~\ref{tab:data_aug}. We find our RGB image augmentation is quite helpful and removing it will hurt the performance. In addition, adding either rotation, scaling, translation or flipping to augment point clouds will hurt the representation learning process. Using the multi-view rendering strategy, we achieves a mIoU of 72.8, which is still not as good as our data augmentation strategy. These validate that our simple data augmentation strategy which involves only data jittering is quite effective empirically. It would be an interesting direction to further explore the best data augmentation strategy as well as the underlying principle for specific downstream tasks in the future.

\begin{table}[h]
\vspace{-5pt}
\centering
\caption{Comparison among different data augmentation strategies. MVR denotes Multi-View Rendering strategy.}
\label{tab:data_aug}
\newcolumntype{Y}{>{\centering\arraybackslash}X}
{
\setlength{\tabcolsep}{0.2em}
\begin{tabularx}{\columnwidth}{>{\centering} m{0.1\columnwidth}|>{\centering} m{0.25\columnwidth}|Y|Y|Y|Y|Y}
\toprule
   Base & No RGB aug & Rot &Scal & Trans & Flip & MVR\\
    \hline
    \textbf{74.6} & 74.1 &70.5 & 71.7 &71.0 &70.4 &72.8 \\
\bottomrule
\end{tabularx}
}
\end{table}

\noindent\textbf{Cross-modal embedding v.s. P4Contrast}
A popular representation learning framework for RGB-D input is to extract features separately from RGB and depth channels and train them with a cross-modal contrasive loss~\cite{meyerimproving,tian2019contrastive,jing2020self,gupta2016cross}. However as we mentioned previously, such approaches usually fail to leverage the synergies between color and geometry inputs to extract features better than using a single modality alone. To demonstrate this, we design a point and pixel level contrastive learning framework using cross-modal contrastive loss and compare with our P4Contrast. To be specific, we use SR-UNet to consume an input point cloud and extract per-point representations. We use FuseNet without the depth fusion layers to consume an input RGB image and extract per-pixel representations. Contrastive loss is then applied between point and pixel representations to bring together corresponding point-pixel pairs and push apart unmatched point-pixel pairs. After pretraining, we concatenate the representations of corresponding point-pixel pairs for downstream tasks. On the ScanNetV2 semantic segmentation task, this cross-modal embedding framework achieves $72.4\%$ mIoU, the same as PointContrast\footnotemark[1] in Table~\ref{tab:scannet_semseg} using only point cloud input and much worse than P4Contrast with a $2.2\%$ mIoU drop. This again justifies our design of P4Contrast and confirms the deficiency of cross-modal embedding as a proper RGB-D representation learning framework.

\noindent\textbf{Fusion architecture study}
With our backbone, we essentially fuse color and geometry signals multiple times. We early fuse color and geometry by combining RGB values and point coordinates together at the input of SR-UNet. We also have a feature level fusion between color and geometry inside FuseNet. By concatenating the output of FuseNet and SR-UNet, we conduct another late fusion. Therefore our backbone can be treated as a hybrid fusion architecture. We compare it with two naive solutions. One is an early fusion strategy with just SR-UNet as the backbone consuming 3D point clouds equipped with color features. The other is a late fusion strategy where we first learn per-point representations from 3D point cloud and learn per-pixel representations from RGB images separately using PointContrast-like frameworks. And then we concatenate the features of corresponding point-pixel pairs as a fused feature representation. To learn per-pixel representations from just RGB images, we simply adapt the FuseNet backbone by removing the depth-fusion layers. Late fusion can not model the cross modal correlations very well and thus only achieves $72.8\%$ mIoU on ScanNetV2, with just $0.6\%$ mIoU improvement compared with the baseline trained from scratch. Jointly considering both color and geometry inputs and contrasting pairs of point-pixel pairs with an early fusion strategy already improves the learned representations a lot and achieves $73.6\%$ mIoU. With the hybrid fusion strategy, we can learn the best multi-modal feature representation, achieving $74.6\%$ mIoU.

\section{Conclusion}
This paper proposes constrasting ``pairs of point-pixel pairs'' as a new method for self-supervised representation learning for points in RGB-D scans. The key idea is to train using hard negatives with disturbed correspondences between RGB and 3D points within the same RGB-D observation, as well as between different observations.  This approach encourages the network to learn a representation that encodes salient features of the RGB, 3D point cloud, and their synergies, which leads to pretrained features that outperform others on two RGB-D scene understanding benchmarks.  This result is very encouraging, and suggests future work investigating its applications to other multi-modal domains.

\section*{Acknowledgement}
We would like to thank Haoqi Yuan, Zihan Jia, and Mingdong Wu for their help in establishing the initial foundation for the project. This work was supported by the Key-Area Research and Development Program of Guangdong Province(2019B121204008) and the Center on Frontiers of Computing Studies (7100602567). 

{\small
\bibliographystyle{ieee_fullname}
\bibliography{main}

\begin{thebibliography}{10}\itemsep=-1pt

\bibitem{asano2020self}
Yuki~M. Asano, Christian Rupprecht, and Andrea Vedaldi.
\newblock Self-labelling via simultaneous clustering and representation
  learning.
\newblock In {\em International Conference on Learning Representations (ICLR)},
  2020.

\bibitem{bachman2019learning}
Philip Bachman, R~Devon Hjelm, and William Buchwalter.
\newblock Learning representations by maximizing mutual information across
  views.
\newblock In {\em Advances in Neural Information Processing Systems}, pages
  15535--15545, 2019.

\bibitem{baltruvsaitis2018multimodal}
Tadas Baltru{\v{s}}aitis, Chaitanya Ahuja, and Louis-Philippe Morency.
\newblock Multimodal machine learning: A survey and taxonomy.
\newblock {\em IEEE transactions on pattern analysis and machine intelligence},
  41(2):423--443, 2018.

\bibitem{bojanowski2017unsupervised}
Piotr Bojanowski and Armand Joulin.
\newblock Unsupervised learning by predicting noise.
\newblock {\em arXiv preprint arXiv:1704.05310}, 2017.

\bibitem{caron2020unsupervised}
Mathilde Caron, Ishan Misra, Julien Mairal, Priya Goyal, Piotr Bojanowski, and
  Armand Joulin.
\newblock Unsupervised learning of visual features by contrasting cluster
  assignments.
\newblock {\em Advances in Neural Information Processing Systems}, 33, 2020.

\bibitem{chen2018progressively}
Hao Chen and Youfu Li.
\newblock Progressively complementarity-aware fusion network for rgb-d salient
  object detection.
\newblock In {\em Proceedings of the IEEE conference on computer vision and
  pattern recognition}, pages 3051--3060, 2018.

\bibitem{chen2020simple}
Ting Chen, Simon Kornblith, Mohammad Norouzi, and Geoffrey Hinton.
\newblock A simple framework for contrastive learning of visual
  representations.
\newblock {\em arXiv preprint arXiv:2002.05709}, 2020.

\bibitem{chen2020big}
Ting Chen, Simon Kornblith, Kevin Swersky, Mohammad Norouzi, and Geoffrey
  Hinton.
\newblock Big self-supervised models are strong semi-supervised learners.
\newblock {\em arXiv preprint arXiv:2006.10029}, 2020.

\bibitem{chen2020improved}
Xinlei Chen, Haoqi Fan, Ross Girshick, and Kaiming He.
\newblock Improved baselines with momentum contrastive learning.
\newblock {\em arXiv preprint arXiv:2003.04297}, 2020.

\bibitem{chen2020mocov2}
Xinlei Chen, Haoqi Fan, Ross Girshick, and Kaiming He.
\newblock Improved baselines with momentum contrastive learning.
\newblock {\em arXiv preprint arXiv:2003.04297}, 2020.

\bibitem{cheng2017locality}
Yanhua Cheng, Rui Cai, Zhiwei Li, Xin Zhao, and Kaiqi Huang.
\newblock Locality-sensitive deconvolution networks with gated fusion for rgb-d
  indoor semantic segmentation.
\newblock In {\em Proceedings of the IEEE conference on computer vision and
  pattern recognition}, pages 3029--3037, 2017.

\bibitem{cheng2020look}
Ying Cheng, Ruize Wang, Zhihao Pan, Rui Feng, and Yuejie Zhang.
\newblock Look, listen, and attend: Co-attention network for self-supervised
  audio-visual representation learning.
\newblock In {\em Proceedings of the 28th ACM International Conference on
  Multimedia}, pages 3884--3892, 2020.

\bibitem{choy20194d}
Christopher Choy, JunYoung Gwak, and Silvio Savarese.
\newblock 4d spatio-temporal convnets: Minkowski convolutional neural networks.
\newblock In {\em Proceedings of the IEEE Conference on Computer Vision and
  Pattern Recognition (CVPR)}, pages 3075--3084, 2019.

\bibitem{couprie2013indoor}
Camille Couprie, Cl{\'e}ment Farabet, Laurent Najman, and Yann LeCun.
\newblock Indoor semantic segmentation using depth information.
\newblock {\em arXiv preprint arXiv:1301.3572}, 2013.

\bibitem{dai2017scannet}
Angela Dai, Angel~X Chang, Manolis Savva, Maciej Halber, Thomas Funkhouser, and
  Matthias Nie{\ss}ner.
\newblock Scannet: Richly-annotated 3d reconstructions of indoor scenes.
\newblock In {\em Proceedings of the IEEE Conference on Computer Vision and
  Pattern Recognition}, pages 5828--5839, 2017.

\bibitem{dai20183dmv}
Angela Dai and Matthias Nie{\ss}ner.
\newblock 3dmv: Joint 3d-multi-view prediction for 3d semantic scene
  segmentation.
\newblock In {\em Proceedings of the European Conference on Computer Vision
  (ECCV)}, pages 452--468, 2018.

\bibitem{d2015review}
Sidney~K D'mello and Jacqueline Kory.
\newblock A review and meta-analysis of multimodal affect detection systems.
\newblock {\em ACM Computing Surveys (CSUR)}, 47(3):1--36, 2015.

\bibitem{gao2019rgb}
Mingliang Gao, Jun Jiang, Guofeng Zou, Vijay John, and Zheng Liu.
\newblock Rgb-d-based object recognition using multimodal convolutional neural
  networks: A survey.
\newblock {\em IEEE Access}, 7:43110--43136, 2019.

\bibitem{garnot2020metric}
Vivien Sainte~Fare Garnot and Loic Landrieu.
\newblock Metric-guided prototype learning.
\newblock {\em arXiv preprint arXiv:2007.03047}, 2020.

\bibitem{grill2020bootstrap}
Jean-Bastien Grill, Florian Strub, Florent Altch{\'e}, Corentin Tallec,
  Pierre~H Richemond, Elena Buchatskaya, Carl Doersch, Bernardo~Avila Pires,
  Zhaohan~Daniel Guo, Mohammad~Gheshlaghi Azar, et~al.
\newblock Bootstrap your own latent: A new approach to self-supervised
  learning.
\newblock {\em arXiv preprint arXiv:2006.07733}, 2020.

\bibitem{gupta2014learning}
Saurabh Gupta, Ross Girshick, Pablo Arbel{\'a}ez, and Jitendra Malik.
\newblock Learning rich features from rgb-d images for object detection and
  segmentation.
\newblock In {\em European conference on computer vision}, pages 345--360.
  Springer, 2014.

\bibitem{gupta2016cross}
Saurabh Gupta, Judy Hoffman, and Jitendra Malik.
\newblock Cross modal distillation for supervision transfer.
\newblock In {\em Proceedings of the IEEE conference on computer vision and
  pattern recognition}, pages 2827--2836, 2016.

\bibitem{hadsell2006dimensionality}
Raia Hadsell, Sumit Chopra, and Yann LeCun.
\newblock Dimensionality reduction by learning an invariant mapping.
\newblock In {\em 2006 IEEE Computer Society Conference on Computer Vision and
  Pattern Recognition (CVPR'06)}, volume~2, pages 1735--1742. IEEE, 2006.

\bibitem{hazirbas2016fusenet}
Caner Hazirbas, Lingni Ma, Csaba Domokos, and Daniel Cremers.
\newblock Fusenet: Incorporating depth into semantic segmentation via
  fusion-based cnn architecture.
\newblock In {\em Asian Conference on Computer Vision (ACCV)}, pages 213--228.
  Springer, 2016.

\bibitem{he2019moco}
Kaiming He, Haoqi Fan, Yuxin Wu, Saining Xie, and Ross Girshick.
\newblock Momentum contrast for unsupervised visual representation learning.
\newblock {\em arXiv preprint arXiv:1911.05722}, 2019.

\bibitem{he2020momentum}
Kaiming He, Haoqi Fan, Yuxin Wu, Saining Xie, and Ross Girshick.
\newblock Momentum contrast for unsupervised visual representation learning.
\newblock In {\em Proceedings of the IEEE/CVF Conference on Computer Vision and
  Pattern Recognition}, pages 9729--9738, 2020.

\bibitem{hou20193d}
Ji Hou, Angela Dai, and Matthias Nie{\ss}ner.
\newblock 3d-sis: 3d semantic instance segmentation of rgb-d scans.
\newblock In {\em Proceedings of the IEEE Conference on Computer Vision and
  Pattern Recognition (CVPR)}, pages 4421--4430, 2019.

\bibitem{huang2019texturenet}
Jingwei Huang, Haotian Zhang, Li Yi, Thomas Funkhouser, Matthias Nie{\ss}ner,
  and Leonidas~J Guibas.
\newblock Texturenet: Consistent local parametrizations for learning from
  high-resolution signals on meshes.
\newblock In {\em Proceedings of the IEEE Conference on Computer Vision and
  Pattern Recognition}, pages 4440--4449, 2019.

\bibitem{jiao2020self}
Jianbo Jiao, Yifan Cai, Mohammad Alsharid, Lior Drukker, Aris~T Papageorghiou,
  and J~Alison Noble.
\newblock Self-supervised contrastive video-speech representation learning for
  ultrasound.
\newblock In {\em International Conference on Medical Image Computing and
  Computer-Assisted Intervention}, pages 534--543. Springer, 2020.

\bibitem{jing2020self}
Longlong Jing, Yucheng Chen, Ling Zhang, Mingyi He, and Yingli Tian.
\newblock Self-supervised modal and view invariant feature learning.
\newblock {\em arXiv preprint arXiv:2005.14169}, 2020.

\bibitem{kundu2020virtual}
Abhijit Kundu, Xiaoqi Yin, Alireza Fathi, David Ross, Brian Brewington, Thomas
  Funkhouser, and Caroline Pantofaru.
\newblock Virtual multi-view fusion for 3d semantic segmentation.
\newblock {\em arXiv preprint arXiv:2007.13138}, 2020.

\bibitem{li2020prototypical}
Junnan Li, Pan Zhou, Caiming Xiong, Richard Socher, and Steven~CH Hoi.
\newblock Prototypical contrastive learning of unsupervised representations.
\newblock {\em arXiv preprint arXiv:2005.04966}, 2020.

\bibitem{liang2018deep}
Ming Liang, Bin Yang, Shenlong Wang, and Raquel Urtasun.
\newblock Deep continuous fusion for multi-sensor 3d object detection.
\newblock In {\em Proceedings of the European Conference on Computer Vision
  (ECCV)}, pages 641--656, 2018.

\bibitem{liang2019strong}
Paul~Pu Liang, Yao~Chong Lim, Yao-Hung~Hubert Tsai, Ruslan Salakhutdinov, and
  Louis-Philippe Morency.
\newblock Strong and simple baselines for multimodal utterance embeddings.
\newblock {\em arXiv preprint arXiv:1906.02125}, 2019.

\bibitem{liang2018multimodal}
Paul~Pu Liang, Amir Zadeh, and Louis-Philippe Morency.
\newblock Multimodal local-global ranking fusion for emotion recognition.
\newblock In {\em Proceedings of the 20th ACM International Conference on
  Multimodal Interaction}, pages 472--476, 2018.

\bibitem{lin2017cascaded}
Di Lin, Guangyong Chen, Daniel Cohen-Or, Pheng-Ann Heng, and Hui Huang.
\newblock Cascaded feature network for semantic segmentation of rgb-d images.
\newblock In {\em Proceedings of the IEEE International Conference on Computer
  Vision}, pages 1311--1319, 2017.

\bibitem{lin2014microsoft}
Tsung-Yi Lin, Michael Maire, Serge Belongie, James Hays, Pietro Perona, Deva
  Ramanan, Piotr Doll{\'a}r, and C~Lawrence Zitnick.
\newblock Microsoft coco: Common objects in context.
\newblock In {\em European conference on computer vision}, pages 740--755.
  Springer, 2014.

\bibitem{liu2018rgb}
Hong Liu, Wenshan Wu, Xiangdong Wang, and Yueliang Qian.
\newblock Rgb-d joint modelling with scene geometric information for indoor
  semantic segmentation.
\newblock {\em Multimedia Tools and Applications}, 77(17):22475--22488, 2018.

\bibitem{long2015fully}
Jonathan Long, Evan Shelhamer, and Trevor Darrell.
\newblock Fully convolutional networks for semantic segmentation.
\newblock In {\em Proceedings of the IEEE Conference on Computer Vision and
  Pattern Recognition (CVPR)}, pages 3431--3440, 2015.

\bibitem{mahendran2018cross}
Aravindh Mahendran, James Thewlis, and Andrea Vedaldi.
\newblock Cross pixel optical-flow similarity for self-supervised learning.
\newblock In {\em Asian Conference on Computer Vision}, pages 99--116.
  Springer, 2018.

\bibitem{meyerimproving}
Johannes Meyer, Andreas Eitel, Thomas Brox, and Wolfram Burgard.
\newblock Improving unimodal object recognition with multimodal contrastive
  learning.
\newblock In {\em IEEE/RSJ International Conference on Intelligent Robots and
  Systems}, 2020.

\bibitem{misra2020self}
Ishan Misra and Laurens van~der Maaten.
\newblock Self-supervised learning of pretext-invariant representations.
\newblock In {\em Proceedings of the IEEE/CVF Conference on Computer Vision and
  Pattern Recognition}, pages 6707--6717, 2020.

\bibitem{morvant2014majority}
Emilie Morvant, Amaury Habrard, and St{\'e}phane Ayache.
\newblock Majority vote of diverse classifiers for late fusion.
\newblock In {\em Joint IAPR International Workshops on Statistical Techniques
  in Pattern Recognition (SPR) and Structural and Syntactic Pattern Recognition
  (SSPR)}, pages 153--162. Springer, 2014.

\bibitem{oord2018representation}
Aaron van~den Oord, Yazhe Li, and Oriol Vinyals.
\newblock Representation learning with contrastive predictive coding.
\newblock {\em arXiv preprint arXiv:1807.03748}, 2018.

\bibitem{park2017rdfnet}
Seong-Jin Park, Ki-Sang Hong, and Seungyong Lee.
\newblock Rdfnet: Rgb-d multi-level residual feature fusion for indoor semantic
  segmentation.
\newblock In {\em Proceedings of the IEEE international conference on computer
  vision}, pages 4980--4989, 2017.

\bibitem{qi2020imvotenet}
Charles~R Qi, Xinlei Chen, Or Litany, and Leonidas~J Guibas.
\newblock Imvotenet: Boosting 3d object detection in point clouds with image
  votes.
\newblock In {\em Proceedings of the IEEE Conference on Computer Vision and
  Pattern Recognition (CVPR)}, pages 4404--4413, 2020.

\bibitem{qi2019deep}
Charles~R Qi, Or Litany, Kaiming He, and Leonidas~J Guibas.
\newblock Deep hough voting for 3d object detection in point clouds.
\newblock In {\em Proceedings of the IEEE International Conference on Computer
  Vision (CVPR)}, pages 9277--9286, 2019.

\bibitem{qi2017pointnet++}
Charles~Ruizhongtai Qi, Li Yi, Hao Su, and Leonidas~J Guibas.
\newblock Pointnet++: Deep hierarchical feature learning on point sets in a
  metric space.
\newblock In {\em Advances in Neural Information Processing Systems (NeurIPS)},
  pages 5099--5108, 2017.

\bibitem{ren2015faster}
Shaoqing Ren, Kaiming He, Ross Girshick, and Jian Sun.
\newblock Faster r-cnn: Towards real-time object detection with region proposal
  networks.
\newblock In {\em Advances in neural information processing systems}, pages
  91--99, 2015.

\bibitem{ronneberger2015u}
Olaf Ronneberger, Philipp Fischer, and Thomas Brox.
\newblock U-net: Convolutional networks for biomedical image segmentation.
\newblock In {\em International Conference on Medical Image Computing and
  Computer-Assisted Intervention (MICCAI)}, pages 234--241. Springer, 2015.

\bibitem{sayed2018cross}
Nawid Sayed, Biagio Brattoli, and Bj{\"o}rn Ommer.
\newblock Cross and learn: Cross-modal self-supervision.
\newblock In {\em German Conference on Pattern Recognition}, pages 228--243.
  Springer, 2018.

\bibitem{shi2020contrastive}
Lei Shi, Kai Shuang, Shijie Geng, Peng Su, Zhengkai Jiang, Peng Gao, Zuohui Fu,
  Gerard de Melo, and Sen Su.
\newblock Contrastive visual-linguistic pretraining.
\newblock {\em arXiv preprint arXiv:2007.13135}, 2020.

\bibitem{shutova2016black}
Ekaterina Shutova, Douwe Kiela, and Jean Maillard.
\newblock Black holes and white rabbits: Metaphor identification with visual
  features.
\newblock In {\em Proceedings of the 2016 Conference of the North American
  Chapter of the Association for Computational Linguistics: Human Language
  Technologies}, pages 160--170, 2016.

\bibitem{song2015sun}
Shuran Song, Samuel~P Lichtenberg, and Jianxiong Xiao.
\newblock Sun rgb-d: A rgb-d scene understanding benchmark suite.
\newblock In {\em Proceedings of the IEEE conference on computer vision and
  pattern recognition}, pages 567--576, 2015.

\bibitem{song2016deep}
Shuran Song and Jianxiong Xiao.
\newblock Deep sliding shapes for amodal 3d object detection in rgb-d images.
\newblock In {\em Proceedings of the IEEE Conference on Computer Vision and
  Pattern Recognition}, pages 808--816, 2016.

\bibitem{thomas2019kpconv}
Hugues Thomas, Charles~R Qi, Jean-Emmanuel Deschaud, Beatriz Marcotegui,
  Fran{\c{c}}ois Goulette, and Leonidas~J Guibas.
\newblock Kpconv: Flexible and deformable convolution for point clouds.
\newblock In {\em Proceedings of the IEEE International Conference on Computer
  Vision}, pages 6411--6420, 2019.

\bibitem{tian2019contrastive}
Yonglong Tian, Dilip Krishnan, and Phillip Isola.
\newblock Contrastive multiview coding.
\newblock {\em arXiv preprint arXiv:1906.05849}, 2019.

\bibitem{wald2019rio}
Johanna Wald, Armen Avetisyan, Nassir Navab, Federico Tombari, and Matthias
  Nie{\ss}ner.
\newblock Rio: 3d object instance re-localization in changing indoor
  environments.
\newblock In {\em Proceedings of the IEEE International Conference on Computer
  Vision}, pages 7658--7667, 2019.

\bibitem{wang2014multi}
Anran Wang, Jiwen Lu, Gang Wang, Jianfei Cai, and Tat-Jen Cham.
\newblock Multi-modal unsupervised feature learning for rgb-d scene labeling.
\newblock In {\em European Conference on Computer Vision}, pages 453--467.
  Springer, 2014.

\bibitem{wang2019densefusion}
Chen Wang, Danfei Xu, Yuke Zhu, Roberto Mart{\'\i}n-Mart{\'\i}n, Cewu Lu, Li
  Fei-Fei, and Silvio Savarese.
\newblock Densefusion: 6d object pose estimation by iterative dense fusion.
\newblock In {\em Proceedings of the IEEE Conference on Computer Vision and
  Pattern Recognition}, pages 3343--3352, 2019.

\bibitem{wu2018unsupervised}
Zhirong Wu, Yuanjun Xiong, Stella~X Yu, and Dahua Lin.
\newblock Unsupervised feature learning via non-parametric instance
  discrimination.
\newblock In {\em Proceedings of the IEEE Conference on Computer Vision and
  Pattern Recognition}, pages 3733--3742, 2018.

\bibitem{xiao2020should}
Tete Xiao, Xiaolong Wang, Alexei~A Efros, and Trevor Darrell.
\newblock What should not be contrastive in contrastive learning.
\newblock {\em arXiv preprint arXiv:2008.05659}, 2020.

\bibitem{xie2020pointcontrast}
Saining Xie, Jiatao Gu, Demi Guo, Charles~R Qi, Leonidas~J Guibas, and Or
  Litany.
\newblock Pointcontrast: Unsupervised pre-training for 3d point cloud
  understanding.
\newblock {\em arXiv preprint arXiv:2007.10985}, 2020.

\bibitem{xu2017multi}
Xiangyang Xu, Yuncheng Li, Gangshan Wu, and Jiebo Luo.
\newblock Multi-modal deep feature learning for rgb-d object detection.
\newblock {\em Pattern Recognition}, 72:300--313, 2017.

\end{thebibliography}
}

\typeout{get arXiv to do 4 passes: Label(s) may have changed. Rerun}
\end{document}